\documentclass[letterpaper, 10 pt, conference]{ieeeconf}  

\IEEEoverridecommandlockouts                              

\usepackage{cite}
\usepackage{amsmath,amssymb,amsfonts}
\usepackage{stackengine}
\stackMath
\usepackage{algorithmic}
\usepackage{graphicx}
\usepackage{textcomp}
\usepackage{xcolor}
\usepackage{tikz}
\usepackage{pgfplots}
\usepackage{subcaption}
\usepackage{booktabs}
\usepackage{adjustbox}
\usepackage{rotating}
\usepackage{geometry}
\usepackage[flushleft]{threeparttable}

\begin{document}

\title{Spring-Brake! Handed Shearing Auxetics Improve Efficiency of Hopping and Standing} 

\author{Joseph Sullivan$^{1,*}$, Ian Sullivan Good$^{2,*}$, Samuel A.\ Burden$^{1}$, Jeffrey Ian Lipton$^{3}$ 
\thanks{$^{*}$ These authors contributed equally to this work}
\thanks{This work was supported by the National Science Foundation, grant numbers 2017927 and 2035717, by the ONR through Grant DB2240 and by the Murdock Charitable Trust through grant 201913596}
\thanks{$^{1}$ Electrical and Computer Engineering, The University of Washington, Seattle, WA 98195 USA}%
\thanks{$^{2}$Mechanical Engineering, University of Washington, Seattle, WA, 98195 USA}%
\thanks{$^{3}$Mechanical and Industrial Engineering, Northeastern University, Boston, MA 02115 USA}
}

\maketitle
\thispagestyle{empty}
\pagestyle{empty}

\begin{abstract}

Energy efficiency is critical to the success of legged robotics. 
Efficiency is lost through wasted energy during locomotion and standing. Including elastic elements has been shown to reduce movement costs, while including breaks can reduce standing costs. 
However, adding separate elements for each increases the mass and complexity of a leg, reducing overall system performance.
Here we present a novel compliant mechanism using a Handed Shearing Auxetic (HSA) that acts as a spring and break in a monopod hopping robot.
The HSA acts as a parallel elastic actuator, reducing electrical power for dynamic hopping and matching the efficiency of state-of-the-art compliant hoppers.
The HSA’s auxetic behavior enables dual functionality.
During static tasks, it locks under large forces with minimal input power by blocking deformation, creating high friction similar to a capstan mechanism. This allows the leg to support heavy loads without motor torque, addressing thermal inefficiency. 
The multi-functional design enhances both dynamic and static performance, offering a versatile solution for robotic applications.

\end{abstract}




\section{Introduction}
\begin{figure}[!htb]
    \centering
    \includegraphics[]{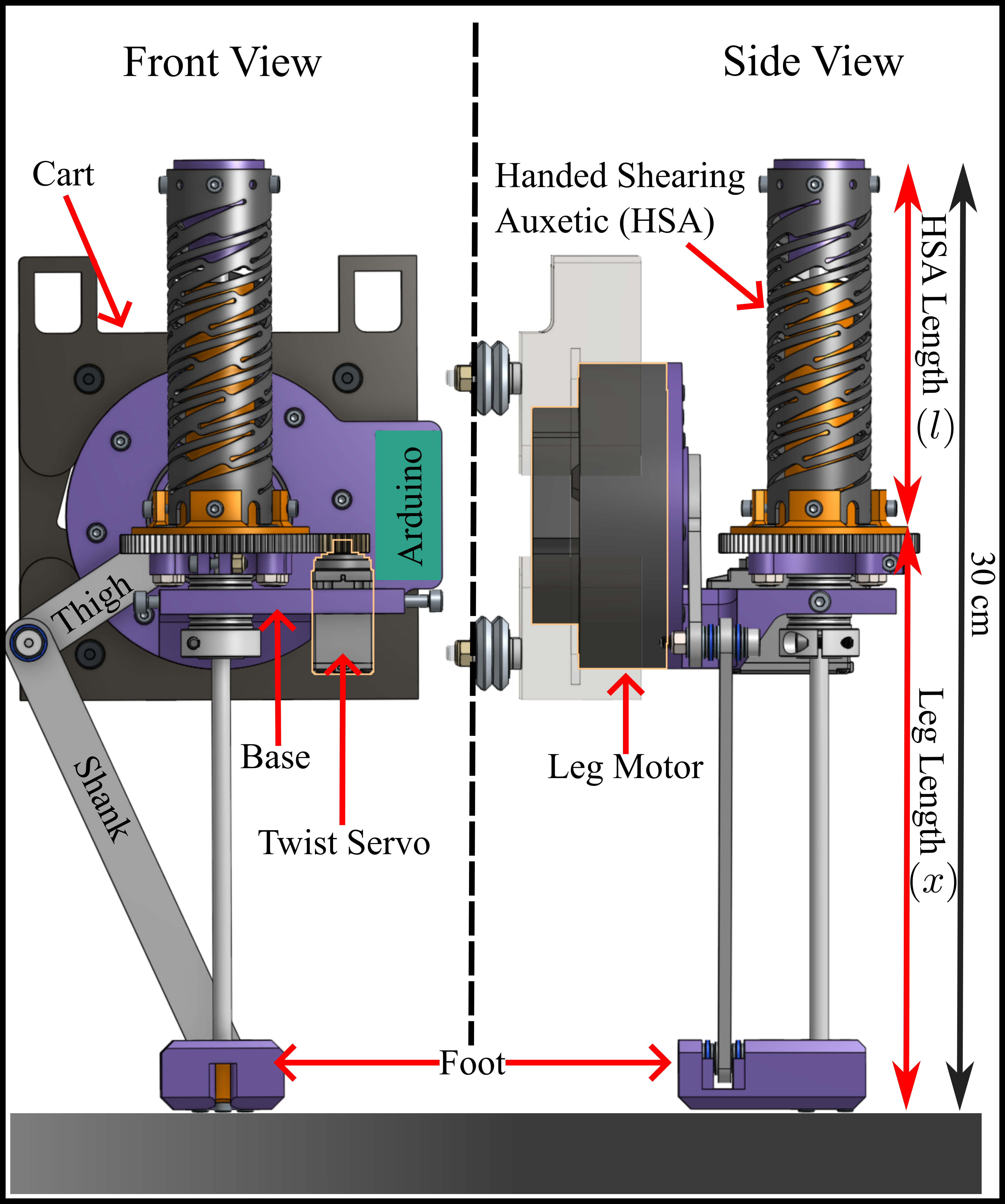}
    \caption{System Overview. We use Handed Shearing Auxetics (HSA) as a combined Spring and Break mechanism in a hopping leg to improve system efficiency in hopping and standing. }
    \label{fig:hero}
    \hfill
\end{figure}

Legged robots have gained traction in human-built environments and unstructured terrain, where wheeled and tracked systems struggle. This makes them well-suited for inspection, surveying, and home assistance, while several companies have also explored their use in logistics. A key limitation is their shorter operating range and endurance, compared to wheeled robots. This is due in part to two problems. One is their higher cost-of-transport (COT) (specific resistance)—a fundamental measure of locomotion efficiency \cite{kashiri2018overview}. Legged robots consistently exhibit higher COT than humans and animals \cite{burden2024animals}, in addition to wheeled machines. Secondly, legged robots often spend energy to maintain position, a critical function in many applications. Achieving greater power autonomy without increasing costs or compromising performance remains a challenge.

%
%

Approaches to reduce cost of transport often focus on improving the electrical efficiency of leg actuators through integration of mechanical compliance \cite{hurst2008role}, locking mechanisms \cite{plooij2015lock}, and regenerative braking \cite{seok2014design}. Mechanical compliance lets joints store and return mechanical energy, reducing the need for motor work. Prominent robots that exploit mechanical compliance include Cassie \cite{reher_dynamic_2019}, ANYmal \cite{hutter2016anymal}, and ATRIAS \cite{hubicki2016atrias}. Locking mechanisms such as cams \cite{Vu2015L-MESTRAN}, ratchets \cite{collins2015reducing}, and clutches\cite{liu2018switchable} have also been employed for redirecting the kinetic energy of swinging legs and in combination with compliance \cite{plooij_clutched_2017}, to decouple springs from joints during specific gait phases \cite{collins2005bipedal}. 

While the role of compliance and locking mechanisms have been extensively studied for dynamic locomotion, less attention has been given to their role in static tasks, such as compensating for gravity across different postures and loads. This functionality is especially important in inspection and surveying applications, where a mobile robot often serves as a stationary sensing platform for extended periods. In these scenarios, wheeled and tracked robots can maintain position indefinitely, whereas most legged robots must continuously consume power to remain in place.

\subsection{Contribution: A Multi-Functional Compliant Mechanism}
In this paper, we present a novel compliant mechanism based on a 3D-printed Handed Shearing Auxetic (HSA) structure, integrated into a monopod hopping robot. The HSA functions as a parallel elastic actuator, providing mechanical compliance alongside a low-reduction motor while enduring large cyclic loads during hopping. We demonstrate that the HSA significantly reduces electrical power consumption during hopping, achieving efficiency comparable to other state-of-the-art compliant hoppers.

Beyond dynamic efficiency, we leverage the auxetic properties of the HSA to enable dual functionality: in static tasks, the HSA acts as a locking mechanism, resisting large forces with minimal input power. This locking effect occurs when the auxetic deformation of the HSA is blocked at high twist angles, generating high friction akin to a capstan mechanism. As a result, the leg adapts its parallel compliance to support heavy loads without requiring continuous torque from the low-reduction motor, which is thermally inefficient under such conditions. This multi-functional design enhances both dynamic performance and static load-bearing capability, offering a versatile solution for robotic actuation.


\section{Background and Related Work}

\subsection{Limiting Factors of Electrical Efficiency in Legged Robots}

Despite their mobility advantages, electrically actuated legged robots suffer from poor energy efficiency compared to wheeled or tracked systems. Key sources of inefficiency include:

\begin{itemize}
    \item \textbf{Joule heating in motors}: leg actuators operate at low speed and high torque during dynamic tasks (e.g. hopping, running), leading to significant thermal losses in electric motors. 
    
    \item \textbf{Inability to recycle energy}: legged locomotion requires a cyclic exchange kinetic and potential energy. This recycling process is least efficient in electric drives at low speed and high torque. During hopping and running, opportunities for recycling energy are further reduced by elastic restitution of velocity at leg impacts.
    
    \item \textbf{static power consumption:} static tasks (e.g. standing under load) demand high sustained torque, which may necessitate high-reduction gearboxes on electric drives or joint locking mechanisms.
\end{itemize}

\subsection{Two Paradigms for Compliant Leg Actuators}

\subsubsection{High Reduction Motors + Series Compliance}
Series elastic actuators (SEAs) are widely used in electrically actuated robots, combining a high-reduction, non-backdrivable motor with a spring placed in series between the motor output and the joint. While high-reduction motors mitigate impact-losses and Joule heating, these efficiencies can be offset by gearbox losses. In \cite{yesilevskiy_comparison_2015}, SEAs have been theoretically shown to be optimal for hopping when regeneration is neglected. However, they introduce challenges such as increased sensor requirements, additional mechanical degrees of freedom, more complex control issues (e.g., stability and phase lag)\cite{kashiri2018overview} and reduced position control bandwidth \cite{hutter2016anymal}.

\subsubsection{Low Reduction Motors + Parallel Compliance}
Parallel elastic actuators (PEAs), though less common, are often utilized in robotics. They employ low-reduction, high-torque motors directly connected to a joint, with mechanical compliance arranged in parallel. This design can significantly enhance energy efficiency by storing mechanical energy and reducing motor torque during cyclic motions. PEAs are a simpler alternative to SEAs, and are well-suited to the dynamics of legged locomotion, which strongly resembles passive spring-mass dynamics \cite{gan2016passive}. Due to reduced gearing, the low-reduction motors in PEAs are more mechanically efficient than their SEA counterparts. However, the key drawback of PEAs are their low thermal efficiency. Additionally, PEAS directly couple the motor to the joint, increasing the joint inertia, resulting in greater impact losses. The need to improve thermal efficiency of direct-drive actuators, while also satisfying torque demands and mitigating impact losses, has led to detailed investigations of optimal design of quasi direct-drive actuators, like those used by Mini Cheetah \cite{hattori2020design}.

\subsection{Handed Shearing Auxetics and Their Application}
Handed Shearing Auxetics (HSAs) are a type of structural metamaterial characterized by their ability to convert shear into volumetric deformations \cite{lipton2018handedness}. The auxetic nature of HSAs means that as a shear stress is applied, not only does the structure elongate, but its diameter increases as well. The linking of shear with extension and diametric expansion allows HSAs to store and release energy in multiple degrees of freedom simultaneously. By placing a smaller diameter object concentric with the HSA, a shear applied to the HSA can cause it to jam against the inner surface, resulting in a non-linear stiffness profile. Because HSAs are size and material invariant, they can be computationally designed to fit a specific application needs. This enables lightweight, compact, and customizable designs that can achieve a broad range of static and dynamic requirements.




HSAs have been used in a wide range of systems from mobile robots \cite{Ketchum2023TurtleGait,kim2024flexible}, to robot grippers \cite{chin2020multiplexed,chen2024real} and even robot arms \cite{garg2022kinematic}. They have been shown to be fabricable from diverse materials: spring steel, PTFE \cite{lipton2018handedness}, and a wide range of printing materials \cite{truby2021recipe,kim2024flexible}. These past works have leveraged the distributed compliance and energy storage of HSAs to sort recycling, walk efficiently, and even activate a can of hair spray using motor imagery brain signals \cite{Chin2019Recycling,Kaarthik2022TurtleBot,Stölzle2024Brain}. However, past works have limited themselves to quasi-static or low frequency operations. This has left HSA systems under explored for highly dynamic tasks.

\section{Methods}
\subsection{Design Principles of an HSA Hopping Robot}
In this paper we set out to design a hopping robot capable of hopping at 5cm height at frequency $f > 2 \; \text{Hz}$. Our approach combines a low-reduction motor with an HSA in parallel. We selected a quasi direct-drive motor for its favorable ratio of torque to reflected inertia and ample regeneration capability.

A prototype robot was first manufactured and weighed. The total mass of the system of 1.3 kg is dominated by the motor and cart cart. To design the HSA, we approximated the robot's stance-phase dynamics with a spring-mass-oscillator model:
$$m \ddot{x} + K(x-x_0) + mg = 0,$$
where $m = 1.3 \; \text{Kg}$ is the total system mass, $x$ represents the leg length, $K$ is the effective stiffness, and $g$ is the gravitational acceleration. 

To compute the required spring constant $K$, we equate the mechanical energy at the lowest point of stance to the potential energy needed for a hop of height $h$, 
$$mgh = \frac{1}{2}K\Delta x^2 - mg\Delta x.$$
Rearranging for $K$,
$$K = 2mg(h+\Delta x)\Delta x^{-2}.$$
Setting $\Delta x = h = 5 \; \text{cm}$ gives an effective spring rate of $K \approx 1 \frac{\text{kN}}{\text{m}}$. 

To verify that this stiffness produces hop frequency of at least 2 Hz, we compute the duration of stance and flight modes:
\begin{equation*}
\begin{array}{cc}
t_F = 2\sqrt{\frac{2h}{g}}, &
t_S(m,K) \approx \pi \sqrt{\frac{m}{K}}
\end{array}
\end{equation*}
The hopping frequency is then:
\begin{equation}
f(m,K) \approx \left(2\sqrt{\frac{2h}{g}}+\pi\sqrt{\frac{m}{K}} \right)^{-1}.
\end{equation}
Substituting $K = 1 \; \frac{\text{kN}}{\text{m}}$, we find:
\begin{align*}
f(1.3,1000)  \approx 3.2 \; \text{Hz}.
\end{align*}

\subsubsection{Final Design Specifications}
From this analysis, we derive the following actuator design requirements:
\begin{itemize}
    \item The actuator \textbf{stroke length} must exceed $\Delta x = 5 \; \text{cm}$
    \item The HSA must provide an \textbf{effective stiffness} of $K \approx 1000 \; \frac{\text{N}}{\text{m}}$
    \item The HSA must support \textbf{loading} at or above: $ 1.5 \times K \times 5 \; \text{cm} = 75 \; \text{N}$
\end{itemize}
\subsection{HSA Selection and Fabrication}

We select an HSA 0\% of the way along the auxetic trajectory as this maximizes the stroke length, minimizing stiffness as shown in \cite{good2022expanding}. Based on work in \cite{good2022expanding,truby2021recipe}, we elect for an HSA with 8 rows and 3 columns as this gives good diametric change for jamming while ensuring we can achieve the stroke length targets. The HSA was made from FPU50 on the Carbon M1 DPL printer following manufactures best practices. FPU50 HSAs can easily support cyclic loads of 50\% strain so the HSA was scaled to a length of 100 mm, with an additional 16 mm added for mounting, resulting in a 3D printed part that is 116mm long and 34mm in diameter. The minimum cross section in the design is $36\;\text{mm}^2$, and with the assumed loading conditions, we see internal stresses of $2.1\;\text{MPa}$, which is well below UTS ($25\;\text{MPa}$).

\begin{figure}[] 
    \centering 
    \includegraphics{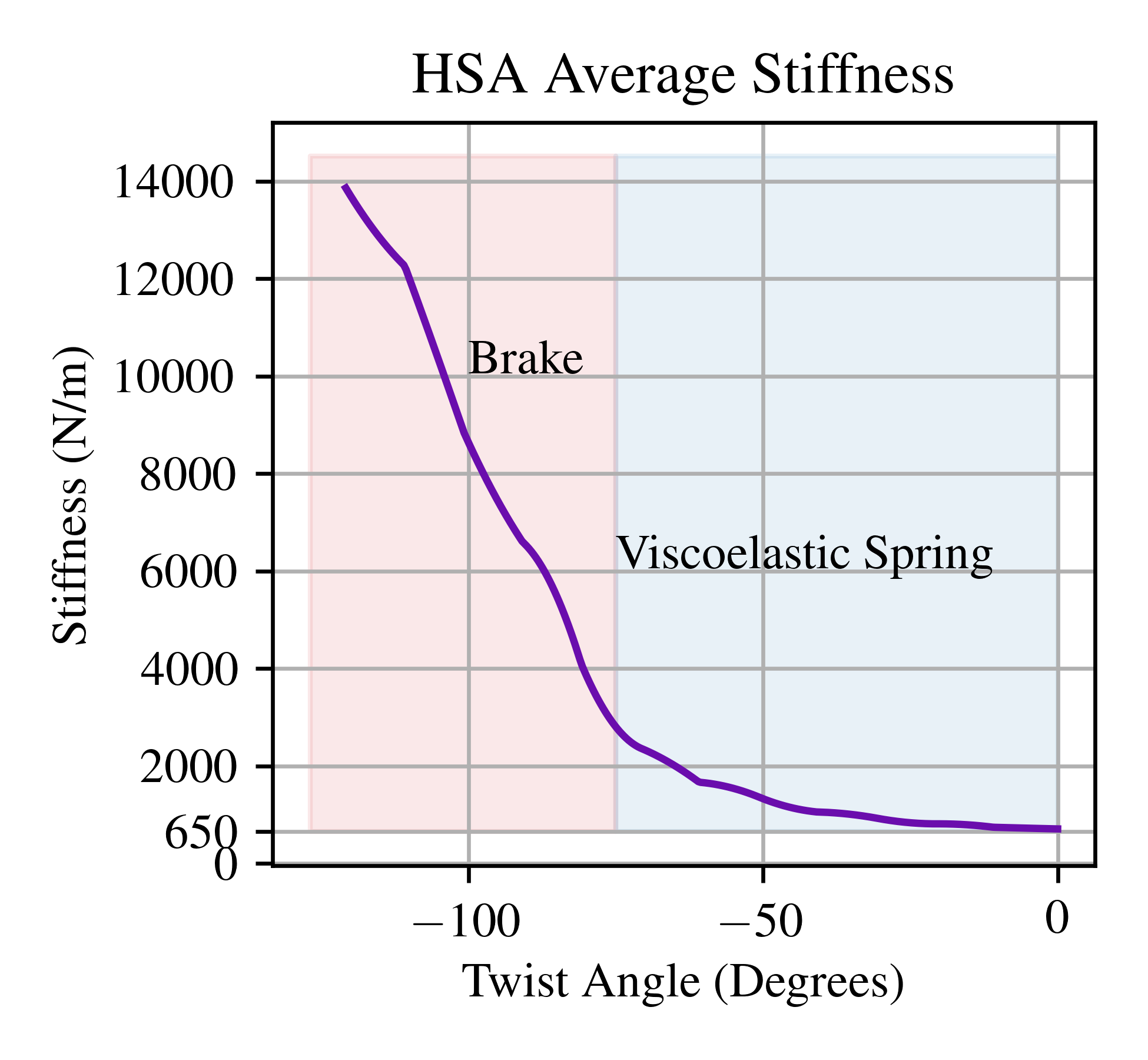} 
    \caption{HSA average stiffness over five-centimeters stroke is shown for different twist angles. Data was collected on an instron UTM. For large twist angles (red region), the HSA jams against the inner cylinder, dramatically increasing its stiffness.} 
    \label{fig:hsa_springBrake}
\end{figure}

This HSA was then characterized on an Instron universal testing machine. A 2D sweep over rotation and displacement was performed. Rotation was swept from $-150^{\circ}$ to $120^{\circ}$ in $10^{\circ}$ increments and displacement was swept from $0$ to $5\text{cm}$ in $5\text{mm}$ increments. The average spring constants from this data are presented in Figure \ref{fig:hsa_springBrake}. Using positive rotation values, the average spring constant was found to be $912 \frac{\text{N}}{\text{m}}$. The HSA demonstrates a peak spring constant over $16 \frac{\text{kN}}{\text{m}}$, a 21$\times$ increase in stiffness compared to the lowest recorded value. This validation confirms that the HSA designed meets or exceeds the final design specifications and will act suitably as both a viscoelastic spring element and as a brake.

\subsection{HSA Parallel Compliance Mechanism}

\begin{figure*}[!htb] \centering \includegraphics{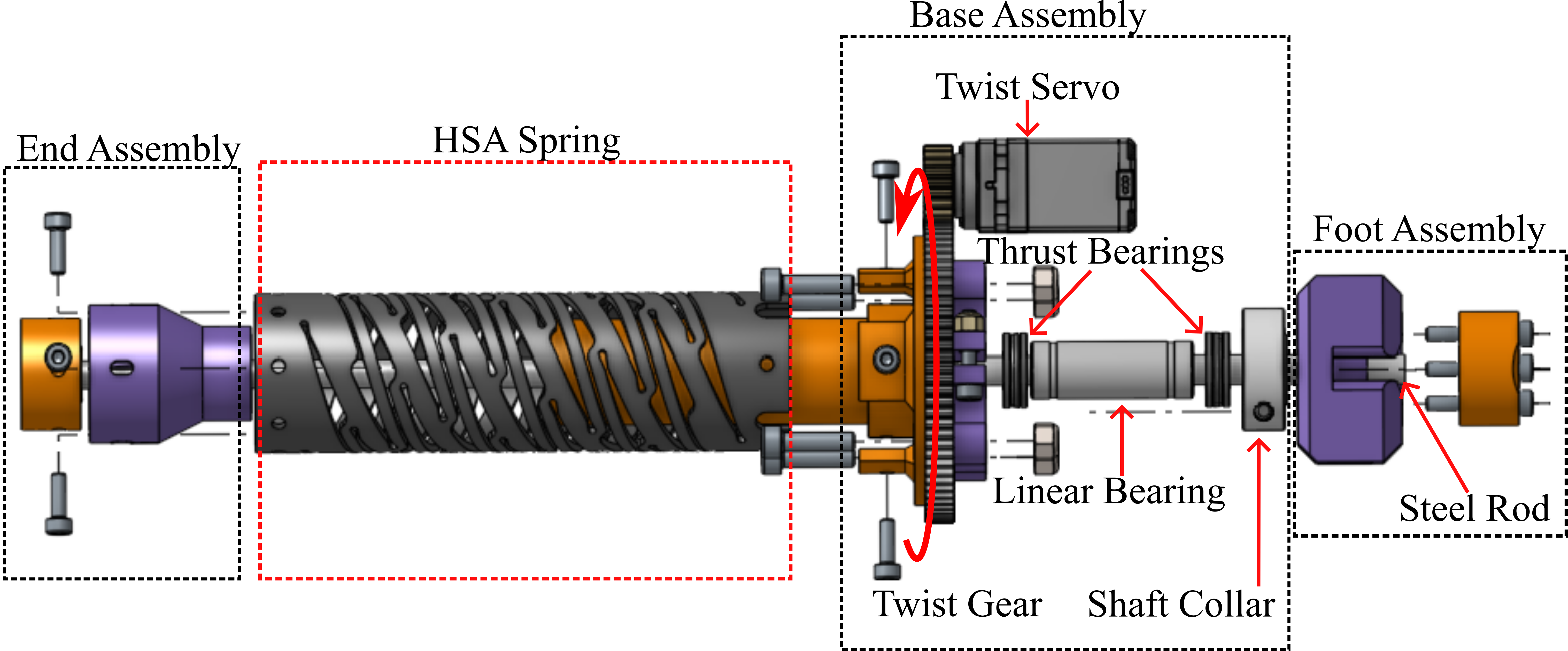} \caption{Exploded view of the HSA assembly, illustrating the twisting mechanism at its base. The HSA is fixed at its right end to a gear carrier, which is driven by a compact servomotor (``Twist Motor"). The left end is coupled to the foot via a 30 cm steel rod passing through a linear bearing embedded in the base. A shaft collar and thrust bearings secure the gear assembly to the cart (not shown). Inside the HSA, a rigid cylindrical insert (occluded in orange) enables the braking mechanism by limiting auxetic deformation.} \label{fig:hsa_expload} \end{figure*}

As shown in Figure \ref{fig:hero}, we integrated an HSA-based compliance mechanism with a prismatic leg actuator to provide parallel elasticity to the leg motor (Mjbots QDD100, with a 1:6 gear reduction). The HSA length $l$ is related to the distance between the foot and body frame $x$ by $l = x - 30\;\text{cm}$. This relationship arises from the closed-loop kinematic structure of the leg, which consists of four primary links (shown in Figure \ref{fig:hero}) :
\begin{enumerate}
\item \textbf{Base link} - Connects the vertical cart to the leg motor and HSA base.
\item \textbf{Thigh link} - Connects the leg motor to the shank.
\item \textbf{Shank link} - Connects the thigh to the foot.
\item \textbf{Foot link} - Connects the shank to the top of the HSA via a steel rod.
\end{enumerate}

\begin{table}
\begin{center}
\label{table:mech_params}
\caption{Robot Mechanical Parameters}
\begin{tabular}{ll}
Thigh  Length & $L_\text{thigh} = 7 \text{cm}$ \\
Shank Length & $L_\text{shank} = 15 \text{cm}$ \\
Rod Length & $L_\text{rod} = 30 \text{cm}$ \\
Cart Mass & $m_c = 1.1 \text{kg}$ \\
Foot Mass & $m_f = 0.2 \text{kg}$ \\ 
Leg Motor Reflected Inerta & $J = 3.5 \times 10^{-3} \text{kg}\text{m}^2$\\
Leg Motor Gear Ratio & $1:6$ \\
Twist Servo Gear Ratio & $1:4$\\
\end{tabular}
\end{center}
\end{table}
\pagebreak
The HSA is mounted to a horizontal gear assembly, positioning the leg motor axis at the HSA's base. The top end of the HSA is fixed to an end assembly that is rigidly attached to the steel rod. At $x = 18.4 $ cm, the HSA reaches its neutral length of 11.6 cm, applying zero force. When $x< 18.4$ cm, the HSA undergoes tensile deformation, generating a reaction force on the leg motor and a corresponding reaction torque on the twist servo.
\subsection{Braking Affordance}
The braking function is enabled by a small servo motor (``Twist Servo'' in Figure \ref{fig:hero}, GoBilda 2000 Series), which actuates the twist gear at the base of the HSA. Rotating this gear counterclockwise deforms the HSA along an auxetic trajectory, causing its inner diameter to contract.

At 135° of twist, this contraction is physically constrained by a rigid cylindrical insert (depicted in orange, occluded in Figure \ref{fig:hero}). Beyond this point, the HSA cannot continue to contract, preventing further deformation. This results in a passive braking effect, in which the HSA remains in a fixed compressed state until the twist servo is reversed.

\begin{table}
\begin{center}
\caption{Leg Motor Electrical Parameters}
\begin{tabular}{ll}
Supply Voltage & $24 \text{V}$ \\
Motor Constant & $105\, \frac{\text{Vs}}{\text{rpm}}$\\
Resistance & $0.143 \, \Omega$\\
Peak Rated Power & $500\, \text{W}$ \\
Peak Rated Torque & $16\, \text{Nm}$
\end{tabular}
\end{center}
\end{table}
\section{Hopping Experiments}
We evaluated the improvement in electrical efficiency provided by the HSA spring through a series of hopping experiments with increasing added mass. These experiments were divided into two groups:
\begin{itemize}
    \item \textbf{With HSA} - The HSA was installed on the actuator (labeled ``with HSA'' in Figure \ref{fig:hop_results}).
    \item \textbf{Without HSA} - The leg motor imposed virtual compliance using proportional angle control (labeled ``without HSA'').
\end{itemize}

Both groups utilized a hybrid PD controller to stabilize the hopping behavior, divided into 'flight' and 'stance' modes. The purpose of the flight control is to lift the foot off the ground, and reposition it in preparation for the following stance phase. To achieve the desired hopping height, a fixed push-off torque is applied in the second half of the stance phase.

During the experiments, we collected telemetry from the leg motor, including angle ($\theta$), torque ($\tau$) and current ($I$), which we used to calculate motor work, joint work and the total electrical losses. In our analysis, we define electrical losses as the sum of net motor work and thermal losses (heat dissipated in the motor windings). Thermal losses in the twist servo were negligible as a result of the high mechanical advantage and friction in its gearbox. The \textbf{electrical power} of the leg motor is approximated by
\begin{equation} \label{eqn:costs}
P_\text{elec} = P_\text{Therm} + P_\text{Motor} = RI^2 + \tau \dot{\theta}.
\end{equation}
To fairly assess the benefits of the HSA, we included electrical regeneration from negative motor work (i.e., regenerative braking) in our analysis. Previous studies of energetics in hopping robots have overlooked electrical regeneration\cite{yesilevskiy_comparison_2015,liu2018switchable,Vu2015L-MESTRAN}, usually due to the complexity of supporting regenerative circuitry. However regenerating power from negative work can significantly lower cost of transport, as demonstrated by the Cheetah robots \cite{Bledt2018Cheeta3,seok2014design}. Fortunately, modern BLDC motor drives with built-in regeneration capabilities—such as the QDD100 motor used in this study—have since become widely available.


\section{Braking Experiment}
To evaluate the power efficiency of static braking, we compared the electrical power consumption of the leg motor (Mjbots QDD100, 1:6 gearing) and the HSA twist motor (GoBilda 2000 Series, 1:4 gearing). The experiment measured the power drawn by each motor while maintaining a fixed posture under increasing applied force force.

The procedure consisted of:
\begin{enumerate}
\item Engaging the HSA braking mechanism, twisting it into its jammed state to block further deformation.
\item Applying an increasing force on the leg using the leg motor, simulating static holding conditions.
\item Measuring the electrical power consumed by the leg motor and the twist motor at different levels of blocked force.
\end{enumerate}
\section{Results}
\subsection{Jammed HSA Reduces Electrical Power Required For Static Braking}
\begin{figure}
    \centering
\includegraphics[width=1\linewidth]{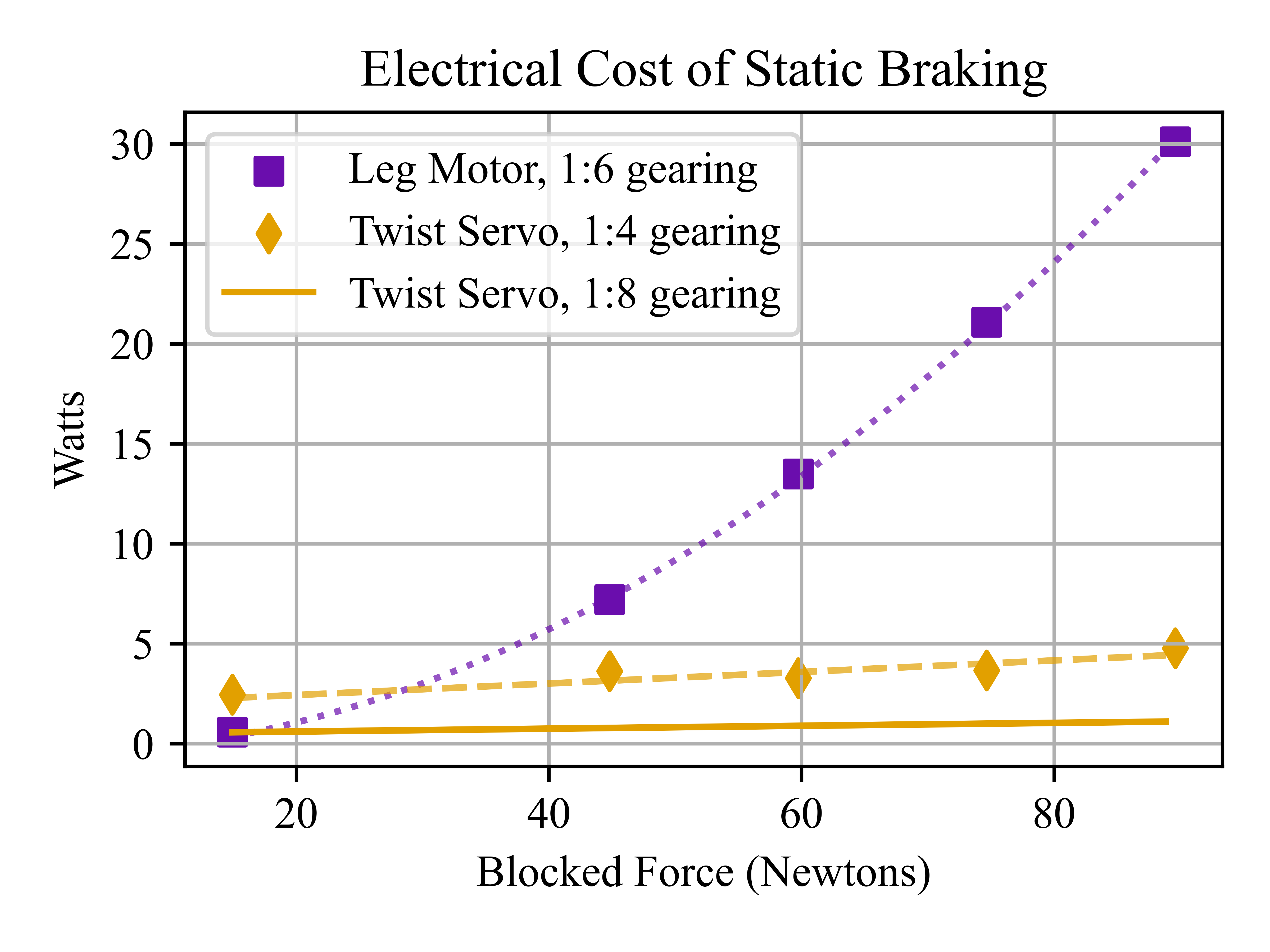}
    \caption{Comparison of the electrical power drawn vs. the force blocked by the leg motor compared to the HSA twist motor. The power consumed by the primary motor (squares) is fit by a quadratic function of the blocked force (purple dotted line). The power consumed by the HSA twist motor (diamonds) follows a linear trend (gold dashed line). The solid gold line predicts the power that would be consumed by the twist servo at 1:8 gear reduction.}
    \label{fig:braking_cost}
\end{figure}
As shown in Figure \ref{fig:braking_cost}, the power consumed by the leg motor increases quadratically with blocked force, while the twist motor follows a linear trend. These results show a jammed HSA can support static loads with significantly lower power consumption than the leg motor motor. A projection for a 1:8 gear reduction (solid gold line) illustrates how additional gearing furter improves the braking efficiency. Theoretically, using a worm gear a jammed HSA can brake for zero electrical cost.

\subsection{3D Printed HSA in Parallel with Low-Reduction Motor Improves Hopping Efficiency}

For each tested mass, between 64 and 77 hops were performed with an HSA and without. The total mean and variance of the hop heights were $5.2 \; \text{cm} \pm 2 \; \text{mm}$. For each condition, we estimated the mean cost-of-tranpsort with 99\% confidence intervals via the boostrapping method.

We found the HSA consistently reduced the hopping cost-of-transport across all tested masses. As stated in section IV, our energy calculation assumes that negative motor work is perfectly regenerated. Without an HSA, the leg motor performs nearly all the negative mechanical work in the system. 


\begin{figure}
    \centering
    \includegraphics[width=1\linewidth]{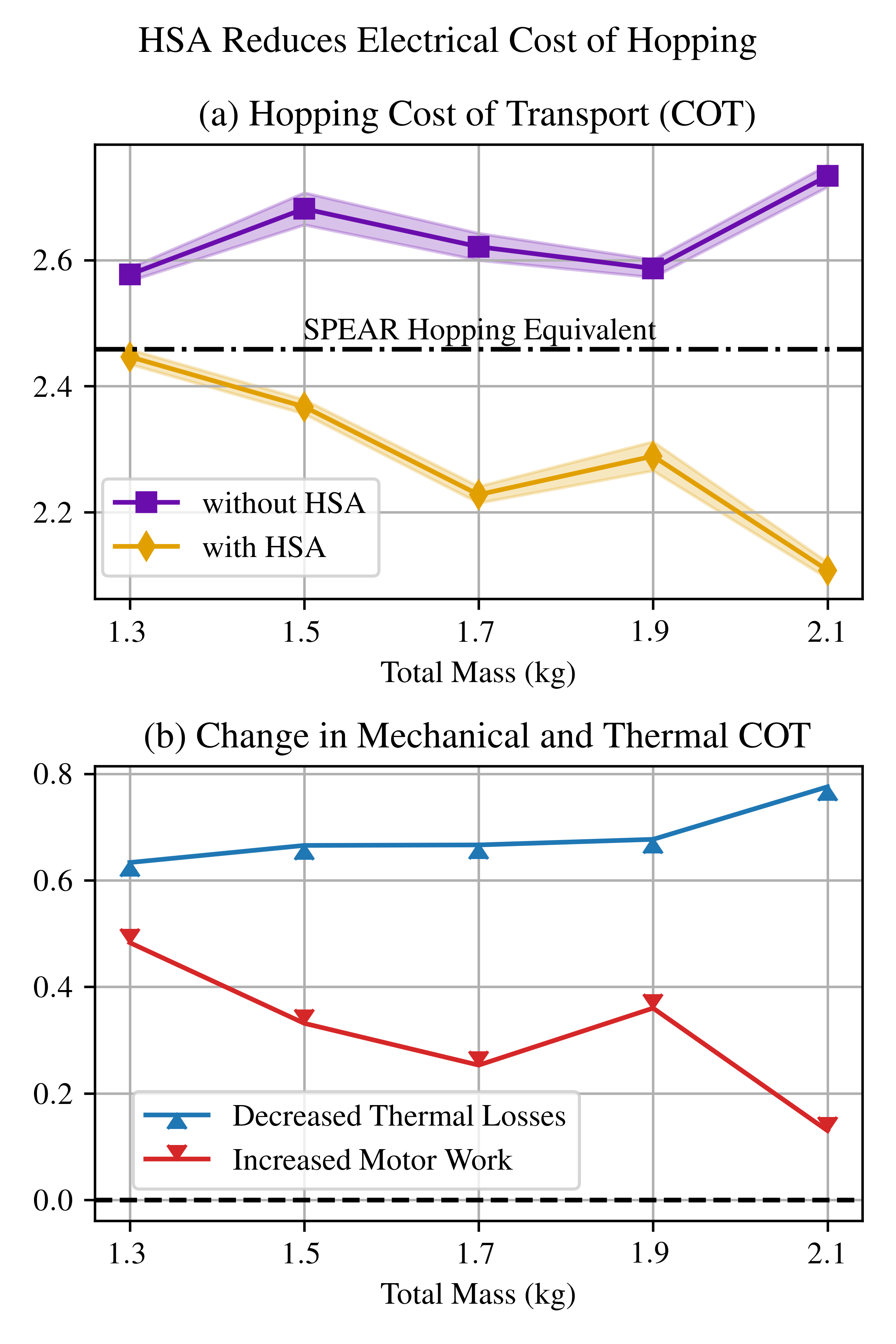}
    \caption{Part (a): Hopping cost-of-transport with and without an HSA compared to results from the SPEAR hopping robot. Part (b): Changes to the mechanical and thermal components of COT (with an HSA minus without an HSA), shown in \emph{absolute value}, i.e. the blue curve is the \emph{decrease} in the thermal cost, the red curve is the \emph{increase} in mechanical cost. The reduced hopping COT in (a) is due to reduced thermal cost, while the mechanical cost increases, suggesting energy is dissipated by the HSA.}
    \label{fig:hop_results}
\end{figure}

\subsection{Increased Efficiency is Driven by Reduced Joule Heating}

Our calculations show an HSA improves the efficiency of hopping by reducing the thermal cost-of-transport (Joule heating), while simultaneously increasing the mechanical cost-of-transport (net motor work).

Figure \ref{fig:hop_results}(b) shows the reduction in the thermal COT due to Joule heating alongside the increase in mechanical COT due to net motor work. These quantities were obtained via numerical integration of the motor and thermal power defined in (\ref{eqn:costs}) during each hop.

\subsection{Cost-of-Transport Comparison with SPEAR Hopper}

To compare the HSA’s performance with traditional steel springs, we analyzed results from the SPEAR hopping robot \cite{liu2018switchable}, where the authors provided the electrical energy $E$ consumed, and force-compression curves $F(x)$ of a \emph{virtual} spring between the robot foot and hip.

We estimated the hop height $h$ by calculating the potential energy stored in the virtual leg spring, and subtracting work against gravity for a lumped mass of 4.91 kg (SPEAR's body weight):
$$h \approx -\Delta x+\frac{1}{mg}\int_0^{\Delta x} F(x)dx .$$

\begin{table}
\begin{threeparttable}
\caption{SPEAR Vertical Hopping Data \cite{liu2018switchable}}
\label{table:spear_data}
\begin{tabular}{llllll}
\toprule
$\theta_\text{knee} \; (\text{rad})$ & $L_\text{td}\; (\text{cm})$ & $E\; (\text{J})$ & $\Delta x\; (\text{cm})$ & $h\; (\text{cm})$ & COT \\ \hline
0.74 & 60.7 & 15 & 10.6 & 10.5 & 2.96 \\
1.04 & 55.7 & 16.3 & 21.8 & 27.5 & 1.23 \\
1.40 & 49.5 & 20 & 27.8 & 34.5 & 1.20\\
\bottomrule
\end{tabular}
\begin{tablenotes}
\item In this table, $\theta_\text{knee}$ is the touchdown angle of the knee joint, $L_\text{td}$ is the length of the virtual leg at touchdown, $E$ is the electrical energy consumed during a hop, $\Delta x$ is the maximum compression of the virtual leg, and $h$ is the calculated hop height.
\end{tablenotes}
\end{threeparttable}
\end{table}

Table \ref{table:spear_data} shows that the COT of SPEAR is linearly related to the hop height ($R^2 = 0.93$). Since these hop heights span a large range of 19\% to 63\% of the virtual leg length, we interpolated the data in Table \ref{table:spear_data} at an intermediate height $h_\text{eq} = 15.8 \; \text{cm}$, calculated by rescaling the average hop height from our data by the ratio of SPEAR's touchdown leg length to our robot's touchdown leg length. This resulted in an equivalent COT of $2.46$, which is shown as a dashed line in Figure \ref{fig:hop_results}(a).


\subsection{3D Printed HSA Dissipates Energy}

Using the data collected during hop experiments, we investigated the amount of positive work that is supplied by the HSA to the system. The positive work done by the HSA can be inferred from the ratio of positive work done by the leg motor to the total positive work in the system. This is captured in the spring efficiency metric:

\begin{equation}
\displaystyle \eta = 1 - \frac{\int_0^{T_S} \max(0,P_\text{Motor}) dt}{\int_0^{T_S} \max(0,P_\text{Joint}) dt}
\end{equation}

where $T_S$ is the duration of the stance period in hopping and $P_\text{Joint}$ is the joint power given by,
\begin{equation*}
P_\text{Joint} = (J\ddot{\theta})\cdot \dot{\theta} + (m_\text{cart}+m_\text{weight}) \left(\ddot{x}+g\right) \cdot \dot{x}.    
\end{equation*}

This efficiency measure was used in \cite{liu2018switchable,hutter2012efficient} to estimate the amount of positive joint work performed by springs in legged robots. In our experiments, we calculated $\eta$ to be $29 \% \;\text{mean} \pm 1.7\; \text{std}\%$, significantly less than the spring efficiency of 64\% reported for SPEAR, suggesting the 3D-printed HSA dissipates more energy than a steel spring.


\section{Conclusion}

In this paper we constructed a hopping robot from a novel combination a 3D printed Handed Shearing Auxetic spring with a low-reduction motor. To the best of our knowledge, this work is the first application of an HSA to dynamic legged locomotion, and the first demonstration of the auxetic jamming effect. 

Our experiments demonstrate that the HSA improves efficiency in both static braking and hopping locomotion. In static conditions, a jammed HSA significantly reduces electrical power consumption compared to a high-torque low-reduction motor, with further improvements possible through additional gearing. During hopping, the HSA consistently reduced electrical cost by 24\%–32\%. This improvement occurred despite reductions in regenerative braking, which is an important efficiency consideration that is often neglected in the legged robotics literature. The primary mechanism behind savings is the reduction of Joule heating in the leg motor. Net motor work increased slightly, but this was outweighed by the reduction in thermal losses, which became more pronounced at higher body masses.

We found that the HSA hopper achieved a cost of transport comparable to SPEAR, a robot using traditional steel springs. However, the spring efficiency $\eta$ was lower, indicating that the 3D-printed HSA dissipates more energy than steel springs, consistent with the increase of net motor work seen in hopping. Nevertheless, our calculations show that the spring provided 35\% of the positive joint work.

Overall the HSA is and effective compliance for direct-drive parallel elastic actuators, where reducing motor losses is crucial.  However, its energy dissipation suggests it may not be as effective in series-elastic configurations, where thermal losses are lower and spring efficiency is more important.

\subsection{Limitations}

The hopping experiments we conducted have two important limitations. First, we only investigated vertical hopping without any horizontal translation, whereas most works involve forward motion of the robot. This discrepancy made a direct comparison of our robot's efficiency to others difficult.

Secondly, in this paper we have not proposed a dynamic model of the HSA, and consequently we did not optimize the leg motor controls for efficiency. A more rigorous investigation into the benefits and drawbacks in hopping locomotion would require a visco-elastic model of the HSA.

\subsection{Future Work}

Our work has shown that the 3D printed HSA is a dissipative element, which limits the maximum efficiency that can be achieved in hopping. This could be improved by increasing the HSA spring efficiency, which may be achieved through alternative materials or manufacturing techniques (e.g., metal HSAs) that may reduce energy dissipation.

The HSA has other properties that are significant to designing leg actuators, but which we did not investigate. In particular, the rest length of an HSA is a nonlinear function of its angular displacement, which has implications for motor Joule heating in quasi-static tasks. Additionally, the effective spring rate of an HSA is nonlinear in both the linear and angular degrees of freedom, a property which may be exploited to dynamically adapt stiffness in different operating conditions.
\section*{Acknowledgment}

Removed in compliance with double-anonymous review requirements.
The authors would like to thank David Oh for his help in designing a dropping mechanism for the leg. Josh Crumrine, and Joe Torky for their help with 3D printing, and Tom Zimet, Jake Madian, and Apoorva Kalaskar, for contributions to earlier versions of robot hardware.

\bibliographystyle{IEEEtran}
\bibliography{hsa}
\end{document}